\documentclass[runningheads]{llncs}
\usepackage{times}
\usepackage{epsfig}
\usepackage{graphicx}
\usepackage{amsmath,amssymb} 
\usepackage{enumitem}
\usepackage{color}
\usepackage{float}
\usepackage{subcaption}

\usepackage{multirow,bigdelim}
\usepackage[pagebackref=true,breaklinks=true,letterpaper=true,colorlinks,bookmarks=false]{hyperref}
\graphicspath{{images/}}

\newcommand{\name}{feature~mapping~RNN~}
\newcommand{\NAME}{Feature~Mapping~RNN~}
\newcommand{\Name}{Feature~mapping~RNN~} 
\newcommand{\etal}{\mbox{\emph{et al.\ }}}
\newcommand{\ie}{\mbox{\emph{i. e.\ }}}
\newcommand{\header}[1]{\noindent \textbf{#1}}

\begin{document}
\title{Action Anticipation with RBF Kernelized \NAME}
\titlerunning{Feature Mapping RNN}
%
\author{Yuge Shi \and
Basura Fernando \and
Richard Hartley}
\authorrunning{Y. Shi et al.}
%
\institute{The Australian National University, Australia}
\maketitle              

\begin{abstract}

We introduce a novel Recurrent Neural Network-based algorithm for future video feature generation and action anticipation called \name. Our novel RNN architecture builds upon three effective principles of machine learning, namely parameter sharing, Radial Basis Function kernels and adversarial training.
Using only some of the earliest frames of a video, the \name is able to generate future features with a fraction of the parameters needed in traditional RNN. 
By feeding these future features into a simple multilayer perceptron facilitated with an RBF kernel layer, we are able to accurately predict the action in the video.

In our experiments, we obtain 18\% improvement on \textit{JHMDB-21} dataset, 6\% on \textit{UCF101-24} and 13\% improvement on \textit{UT-Interaction} datasets over prior state-of-the-art for action anticipation.
\end{abstract}

\section{Introduction}\label{sec:introduction}

\begin{figure}[t]
      \centering
      \includegraphics[width = 0.7\textwidth]{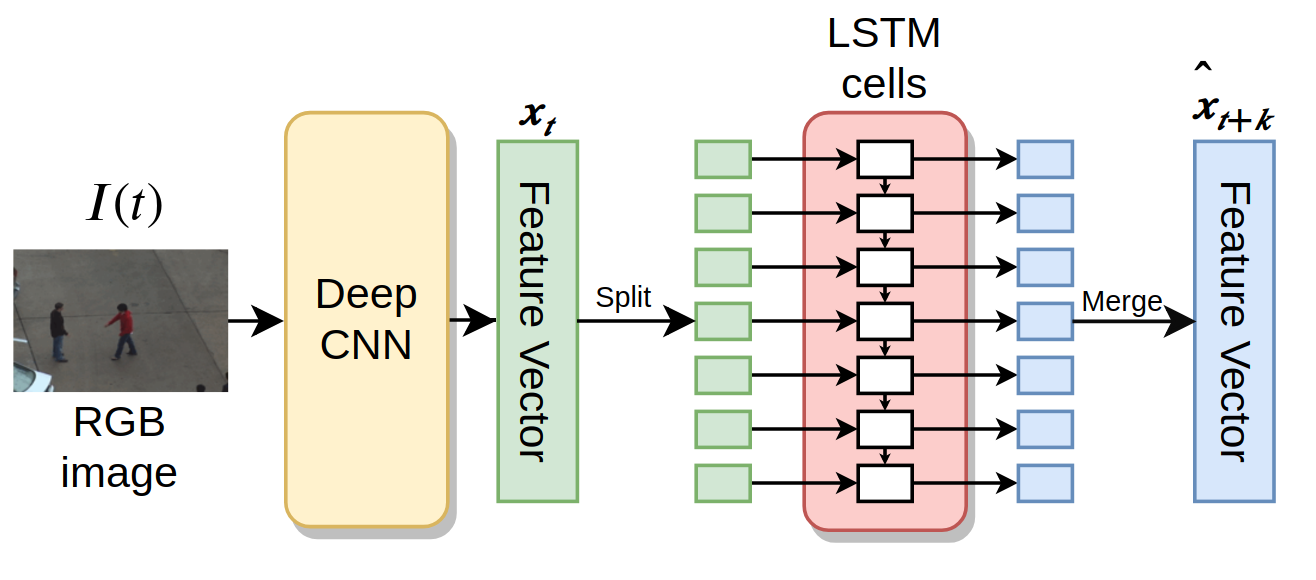}
      \caption{\textbf{Overview of Proposed \name: }Given a frame extracted from video data, the algorithm first passes the RGB image  $I(t)$ through a deep CNN to acquire high level features of the image $\boldsymbol{x_t}$. The vector is then split into smaller segments $\boldsymbol{x_t^i}$ of equal length. Each scalar element in the segmented vector is used as input to a single LSTM cell that produces the prediction of corresponding feature element in frame $(t+k)$, where $k\geq 1$. After all segments are processed with LSTMs, all the prediction segments $\boldsymbol{\hat{x}^i_{t+k}}$ are concatenated back together to form $\boldsymbol{\hat{x}_{t+k}}$, which contains high level features of $I(t+k)$. }\label{fig:opvga} 
\end{figure} 

Action anticipation (sometimes referred to as action prediction) is gaining a lot of attention due to its many real world applications such as
human-computer interaction~\cite{hcinteraction,hcinteraction1,hcinteraction2}, sports analysis~\cite{sports,sports1,sports2} and pedestrian movement prediction~\cite{p1,p2,p3,p4,p5} especially in the autonomous driving scenarios. \par
In contrast to most widely studied human action recognition methods, in action anticipation, we aim to recognize human action as early as possible~\cite{AkbarianSSFPA17,hemakoppula2009,shugaoma2016,soomro2016online,carlvondrick2016}. This is a challenging task due to the complex nature of video data. Although a video containing a human action consists of a large number of frames, many of them are not representative of the action being performed; large amount of visual data also tend to contain entangled information about variations in camera position, background, relative movements and occlusions. This results in cluttered temporal information and makes recognition of the human action a lot harder. The issue becomes even more significant for action anticipation methods, as the algorithm has to make a decision using only a fraction of the video at the very start. Therefore, finding a good video representation that extracts temporal information relevant to human action is crucial for the anticipation model.\par

To over come some of these issues, we resort to use deep convolutional neural networks (CNNs) and take the deep feature on the penultimate layer of CNN as video representation. Another motivation to use deep CNNs stems from the difficulty of generating visual appearances for future. Therefore, similar to Vondrick~\etal~\cite{carlvondrick2016}, we propose a method to generate future features tailored for action anticipation task: given an observed sequence of deep CNN features, a novel Recurrent Neural Network (RNN) model is used to generate the most plausible future features and thereby predicting the action depicted in video data. An overview of this model can be found in Fig.~\ref{fig:opvga}.

The objective of our RNN is to map the feature vector at time $t$ denoted by $\boldsymbol{x_t}$ to the future feature vector at $(t+k)$ denoted by $\boldsymbol{x_{t+k}}$. Because only a fraction of the frames are observed during inference, the future feature generator should be highly regularized to avoid over-fitting. Furthermore, feature generator needs to model complex dynamics of future frame features. \par
This can be resolved by parameter sharing. Parameter sharing is a strong machine learning concept that is being used by many modern leaning methods. Typically, CNNs share parameters in the spatial domain and RNNs in the temporal dimension. In our work, we propose to utilize parameter sharing in an unconventional way for RNN models by expanding it to the feature domain. This is based on the intuition that the CNN feature activations are correlated to each other. \par
By utilizing parameter sharing across feature activations, our proposed RNN is able to learn the temporal mapping from $\boldsymbol{x_t}$ to $\boldsymbol{x_{t+k}}$ with significantly fewer parameters. This greatly boosts the computational efficiency of the prediction model and correspondingly shortens the response time. 
We call our novel RNN architecture \name.\par

To model complex dynamic nature of video data, we make use of a novel mapping layer inside our RNN. 
In principle, the hidden state of the RNN captures the temporal information of observed sequence data. In our method, hidden state of the RNN is processed by a linear combination of Gaussian Radial Basis Function (RBF) kernels to produce the future feature vector. 
While a linear model defines a simple hyperplane as mapping functions, the kernelized mapping with RBF kernels can model complex surfaces and therefore has the potential of improving the prediction accuracy. 
In our work, we also implement RBF kernels on the action classification multi-layer perceptron to improve the performance of classifiers. \par

Ideally, we are interested in learning the probability distribution of future given the past features. To learn this conditional distribution, inspired by the of Generative Adversarial Networks~\cite{gan}, an adversarial approach is used to evaluate the cost of the feature mapping RNN. The RNN is trained with an adversarial loss and re-constrictive L2 loss. 
In this way, the model is optimized not only with the intention of reducing the Euclidean distance between the prediction and ground truth, but also taking probability distribution of the feature vector into consideration.\par

In a summary, our contributions are:
\vspace{-7pt}
\begin{itemize}[leftmargin=.5in]
 \item We propose a novel RNN architecture that share parameters across temporal domain as well as feature space.
 \item We propose a novel RBF kernel to improve the prediction performance of RNNs.
 \item We demonstrate the effectiveness of our method for action anticipation task beating state-of-the-art on standard benchmarks.
\end{itemize}

\section{Related Work}\label{sec:relatedwork}

The model proposed in this paper focuses on future video content generation for action prediction and action anticipation~\cite{hemakoppula2009,tuanhungwu2014,msryoo2009,msryoo2011,yu2012predicting,Li2014,kong2014discriminative,soomro2016online,AkbarianSSFPA17,ma2016learning,carlvondrick2016,gao2017red}.
In contrast to the widely studied action recognition problem, the action anticipation literature focuses on developing novel loss functions to reduce the predictive generalization error~\cite{AkbarianSSFPA17,ma2016learning,jain2016} or to improve the generalization capacity of future content such as future appearance~\cite{gao2017red} and future features~\cite{carlvondrick2016}.
The method propose in this paper also focuses on future content generation and therefore could further benefit from novel loss functions as proposed in~\cite{AkbarianSSFPA17,ma2016learning,jain2016}.

In the early days, Yu~\etal~\cite{yu2012predicting} make use of spatial temporal action matching to tackle early action prediction. Their method relies on spatial-temporal implicit shape models.
By explicitly considering all history of observed features, temporal evolution of human actions is used to predict the class label as early as possible by Kong~\etal~\cite{kong2014discriminative}.
Li~\etal's work~\cite{Li2014} exploits sequence mining, where a series of actions and object co-occurrences are encoded as symbolic sequences. Soomro~\etal~\cite{soomro2016predicting} propose to use binary SVMs to localize and classify video snippets into sub-action categories, and obtain the final class label in an online manner using dynamic programming.
In~\cite{tuanhungwu2014}, action prediction is approached using still images with action-scene correlations. Different from the above mentioned methods, our work is focused on action anticipation from videos.
We rely on deep CNNs along with a RNN that shares parameters across both feature and time dimensions to generate future features. To model complex dynamics of video data, we are the first to make use of effective RBF kernel functions inside RNNs for the action anticipation task.

On the other hand, feature generation has been studied with the aim of learning video representation, instead of specifically for action anticipation. Inspired by natural language processing technique~\cite{Bengio2003}, authors in~\cite{ranzato2014} propose to predict the missing frame or extrapolate future frames from an input video sequence. However, they demonstrate this only for unsupervised video feature leaning. Other popular models include the unsupervised encoder-decoder scheme introduced by~ \cite{SrivastavaMS15} for action classification, probabilistic distribution generation model by~\cite{christophlampert2015} as well as scene prediction learning using object location and attribute information introduced by~\cite{davidfouhey2014}. Research in recent years on applications of Generative Adversarial Network on video generation have given rise to models such as MoCoGAN~\cite{SergeyTulyakov}, TGAN~\cite{MasakiSaito} and Walker~\etal's work~\cite{JacobWalker} on video generation using pose as a conditional information. The mechanisms of these GAN variations are all capable of exploiting both the spatial and temporal information in videos, and therefore have showed promising results in video generation.

Moreover, trajectory prediction~\cite{julianKooij2014}, optical-flow prediction~\cite{jacobwalker2015}, path prediction~\cite{jacobwalker2014,danxie2013} and motion planning~\cite{haifenggong2011,krisKitani2012}, sports forecasting~\cite{PannaFelsen}, activity forecasting of~\cite{TahmidaMahmud} are also related to our work. All these methods generate future aspects of the data. Our novel RNN model, however, focuses on generating future features for action anticipation.



\section{Approach}\label{sec:ourapproach}
\subsection{Overview}
Similar to methods adopted by other action anticipation algorithms, our algorithm makes predictions of action by only observing a fraction of video frames at the beginning of a long video.
The overall pipeline of our method is shown in Fig.~\ref{fig:opvga}.
First, we extract some CNN feature vectors from frames and predict the future features based on the past features. 
Subsequently, a multilayer perceptron (MLP) is used to classify generated features. 
We aggregate predictions from observed and generated features to recognize the action as early as possible.

\subsection{Motivation}
Denote observed sequence of feature vectors up to time $t$ by $X = \left< \boldsymbol{x_1, x_2, x_3, \cdots x_t } \right>$ and future feature vector we aim to produce by $\boldsymbol{\hat{x}_{t+k}}$, where $k\ge1$ and $\boldsymbol{x_t} \in \mathbb{R}^d$. 
We are interested in modeling the conditional probability distribution of $P(\boldsymbol{x_{t+k}} |$ $\boldsymbol{x_1,}$ $\boldsymbol{x_2,}$ $\boldsymbol{x_3,}$ $\boldsymbol{\cdots}$ $\boldsymbol{x_t};\Theta)$, where $\Theta$ denotes the parameters of the probabilistic model. \par
It is natural to use RNNs or RNN variants such as Long Short Term Memory (LSTM)~\cite{hochreiter1997long} to model the temporal evolution of the data.
However, learning such a mapping could lead to over-fitting since these methods tend not to utilise the temporal coherence and the evolutionary nature of video data~\cite{Mobahi2009}.

Furthermore, a naive CNN feature mapping using a LSTM from past to the future is also prone to over-fitting. 
A LSTM with hidden state of dimensionality $H$ and takes feature vectors of dimensionality $d$ as input uses parameters in the order of $4 (dH+d^2)$. As an example, if we use the penultimate activations of Inception V3~\cite{Szegedy2016} as feature vectors ($d=2048$), a typical LSTM ($H=512$) would require parameters in the order of $10^7$.
We believe that the effectiveness of such models can be largely improved by utilising the correlation of high level activations of modern CNN architectures~\cite{Szegedy2016,2015kaimingheRESENET}.

Motivated by these arguments, we propose to train a LSTM model where parameters are not only shared in the time domain, but also across feature activations.
By doing so, we aim to self-regularize the feature generation of the algorithm. We name our novel architecture \emph{\name}.
Furthermore, to increase the functional capacity of RNNs, we make use of Radial Basis Functions (RBF) to model temporal dynamics of the conditional probability distribution $P(\boldsymbol{x_{t+k}}$ $|$ $\boldsymbol{x_1,}$ $\boldsymbol{x_2,}$ $\boldsymbol{x_3,}$ $\boldsymbol{ \cdots}$ $\boldsymbol{x_t};$ $\Theta)$. These mechanisms will be introduced in details in the following subsection.

\subsection{\NAME with RBF Kernel Mapping}
A traditional feature generation RNN architecture takes a sequence of vectors up to time $t$ as input and predicts the future feature vector $\boldsymbol{\hat{x}_{t+k}}$. Typically, the following recurrent formula is used to model the prediction:
\begin{equation}
 \boldsymbol{h_t} = f(\boldsymbol{x_t},\boldsymbol{h_{t-1}};\theta)
 \label{eq.rnn}
\end{equation}
Where $\boldsymbol{h_t}$ is the hidden state ($\boldsymbol{h_t} \in \mathbb{R}^{H}$) which captures the temporal information of the sequence and $\theta$ are the parameters of the recurrent formula. Then we utilize this hidden state to predict the future feature vector $\boldsymbol{x_{t+k}}$ using the following formula:
\begin{equation}
 \boldsymbol{\hat{x}_{t+k}} = \boldsymbol{h_t} \times \boldsymbol{W}
 \label{eq.rnnprd}
\end{equation}
where $\boldsymbol{W} \in \mathbb{R}^{H \times D}$ is the parameter that does the linear mapping to predict the future feature vector.

As introduced previously, in our \name the parameters $\Theta$ are shared across several groups of feature activations. This is achieved by segmenting the input feature vector of dimensionality $d$ into equal size sub-vectors of dimensionality $D$, where $D$ is referred to as \emph{feature step size}.

Now let us denote the \text{$i^{th}$} sub-feature vector of size $D$ by $\boldsymbol{x^{i}_t}$. Intuitively, if we concatenate all such sub-feature vectors in an end-to-end manner, we will be able to reconstruct the original feature vector $\boldsymbol{x_t}$.  The time sequence of data for the \text{$i^{th}$} sub-feature vector is now denoted by $X^{i}=\left<\right.\boldsymbol{x^{i}_1, x^{i}_2,}$ $\boldsymbol{x^{i}_3, \cdots x^{i}_t}\left.\right>$.
If we process each sequence $X^{i}$ in units of $\boldsymbol{x^{i}_t}$ with the RNN model in equation~\ref{eq.rnn} and equation~\ref{eq.rnnprd}, we will be able to predict $\boldsymbol{x^{i}_{t+k}}$ and by concatenating them end-to-end, generate $\boldsymbol{x_{t+k}}$. This approach reduces the number of parameters used in the RNN model from $4 (dH+d^2)$ to $4 (DH+D^2)$, which results in a considerable boost in computational efficiency especially when $D\ll d$.
However, the parameter complexity of the model would remain polynomial and is relevant to multiple hyperparameters.

To further improve the efficiency of our model, we adopt an even bolder approach: we propose to convert the sequence of vectors of 
$X^{i}=\left<\right.\boldsymbol{x^{i}_1, x^{i}_2,}$ $\boldsymbol{x^{i}_3, \cdots x^{i}_t}\left.\right>$ to a sequence of scalars. 
Let us denote the \textit{j-th} dimension of sub-vector $\boldsymbol{x^{i}_t}$ by $x^{i(j)}_t$.
Now instead processing sequence of vectors $X^{i}$, we convert the sequence $X^{i}$ to a new sequence of scalars $X^{'}{^{i}} = \left<\right. x^{i(1)}_1, x^{i(2)}_1, \cdots x^{i(D)}_1, x^{i(1)}_2, x^{i(2)}_2, \cdots, x^{i(k)}_t,$ $\cdots x^{i(D)}_t \left.\right>$. 
Size of the sequence of scalars $X^{'}{^{i}}$ is equal to $t\times D$ and we generate $\frac{d}{D}$ number of such sequences from each original sequence of feature vector $X$.

We then propose to process sequence of scalars using a RNN (LSTM) model. The computation complexity is now linear, with number of parameters used in the recurrent model (LSTM) reduced to $4(H+1)$ and depends only on the hidden state size.

Again, given the current sequence of vectors $X$, we want to generate future feature vector $\boldsymbol{x_{t+k}}$.
In the our RNN model, this is translated to predicting sequence of scalars $\left< x^{i(1)}_{t+k}, \cdots x^{i(D)}_{t+k} \right>$ from sequence $X^{'}{^{i}}$ for all sub-feature vectors $i=1$ to $\frac{d}{D}$.
Then we merge all predicted scalars for time $t+k$ to obtain $\boldsymbol{x_{t+k}}$.

Therefore, mathematically our new RNN model that share the parameter over feature activations can be denoted by the following formula:
\begin{equation}
 \boldsymbol{h^{i(l)}_{t}} = f( x^{i(l)}_{t} , \boldsymbol{h^{i(l)}}_{t-1} ;\Theta^{'})
\end{equation}
where $\Theta^{'}$ is the new parameter set of the RNN (LSTM) and the future \textit{l-th} scalar of \textit{i-th} sub-feature vector is given by:
\begin{equation}
 \hat{x}^{i(l)}_{t+k} = \boldsymbol{h^{i(l)}_{t} \boldsymbol{\cdot} w^{'}}.
\end{equation}
To further improve the functional capacity of our \name, we make use of Radial Basis Functions (RBF). 
Instead of using a simple linear projection of the hidden state to the future feature vector, we propose to exploit the more capable Radial Basis Functional mapping. 
We call this novel RNN architecture the RBF kernelized \name, denoted by the following formula:
\begin{align} 
\label{eq:ksvm}
 \hat{x}^{i(l)}_{t+k} =  \displaystyle\sum_{j=1}^{n}\alpha_j^{l}\ \text{exp}\Big[-\frac{(\boldsymbol{h^{i(l)}_t}-\boldsymbol{\mu_j^{l}})^2}{{\sigma^{l}_j}^{2}}\Big]  
\end{align}
where $\boldsymbol{\mu_j^{l}}$, $\sigma_j^{l}$ and $\alpha_j^{l}$ are parameters learned during training and $n$ the number of RBF kernels used. 
These parameters are shared across all sub-feature vectors.
The future feature vector $\boldsymbol{\hat{x}^{i}_{t+k}}$ is calculated as the linear combination of RBF kernels outputs. Since the RBF kernels are better at modeling complex planes in the feature space, this functional mapping is able to accurately capture more complicated dynamics. Implementing the kernalised RBF on our \name enables the model to do so with fewer parameters than classical RNNs.
Illustration of our RBF kernelized MLP is shown in Fig~\ref{fig:kMLP}.

Note that the method we have presented here only uses non-overlapping feature-sub-vectors, \ie no overlapping exists between 2 consecutive sub-vectors. However, overlapping feature-sub-vectors can be used to improve the robustness of feature generation. Therefore, instead of using a non-overlapping \emph{feature stride} of $D$, we use an overlapping stride of size $S$. In this case, we take the average between all overlapping parts of 2 consecutive sub-vectors to obtain $\hat{x}^{i(l)}_{t+k}$. 

\subsection{Training of feature mapping RNN}

Data generation, especially visual data generation with raw images, has remained a challenging problem for years mainly due to the absence of suitable loss function. 
The most commonly used function for this task is the $\mathcal{L}_2$ loss. However, it works under the assumption that data is drawn from a Gaussian distribution, which makes the loss function ineffective when dealing with data that follows other distributions. As an example, if there exists only two equally possible value $v_1$ and $v_2$ for a pixel, the possibility for $v_{avg}=(v_1+v_2)/2$ to be the true value for that pixel is minimal. However, $v_{avg}$ will be assigned to the output in a neural network that uses $\mathcal{L}_2$ loss to evaluate the cost. This property of the $\mathcal{L}_2$ loss function causes a "blurry" effect on the generated output. Similar observations can be seen for feature vector generation.\par 

Recent developments in Generative Adversarial Networks address this issue successfully~\cite{gan}. 
Traditional GAN consists of 2 CNNs, one of them is named \textit{generator} (denote as $\mathcal{G}$) and the other \textit{discriminator} (denote as $\mathcal{D}$). The GAN effectively learns the probabilistic distribution of the original data, and therefore eliminates the "blockiness" effect caused by $\mathcal{L}_2$ loss function. Here, we propose to train the \name algorithm using a combination of $\mathcal{L}_2$ and adversarial loss, which is realized by implementing the \name as the \textit{generator} denoted by  $\mathcal{G}: \boldsymbol{x_t^{i}}\rightarrow \boldsymbol{\hat{x}_{t+k}^{i}}$. 
By doing so, we are able to produce prediction that is both accurate and realistic.

\emph{${\mathcal{L}_2}$ loss:} The $\mathcal{L}_2$ loss is defined as the mean squared error between the generated feature and the real feature vector of the future frame given as follows:
\begin{equation}
\mathcal{L}_{2}^{\mathcal{G}}(\boldsymbol{x_t}) = ||\boldsymbol{x^{i}_{t+k}}-\boldsymbol{\hat{x}^{i}_{t+k}}|| =  ||\boldsymbol{x^{i}_{t+k}}-\mathcal{G}(\boldsymbol{x^{i}_{t}}) ||. 
\label{eq:l2loss}
\end{equation}

\emph{Adversarial loss:} We use generator adversarial loss proposed by~\cite{gan} where we train $\mathcal{G}$ so that $\mathcal{D}$ believes $\mathcal{G}(\boldsymbol{x_t^{i}})$ comes from the dataset, at which point $\mathcal{D}(\mathcal{G}(\boldsymbol{x_t^i})) = 1$. The loss function is defined as:
\begin{eqnarray}
\mathcal{L}^{\mathcal{G}}_{adv} = -\log(\mathcal{D}(\mathcal{G}(\boldsymbol{x_{t}^{i}}))).
\label{eq:lgadv}
\end{eqnarray}
By adding this loss to our objective function, the RNN is encouraged to generate feature prediction with probabilistic distribution similar to the original data. Finally, the loss function of our RNN generator $\mathcal{G}$ is given by:
\begin{equation}
\mathcal{L}^{\mathcal{G}} = \lambda_1\mathcal{L}^{\mathcal{G}}_{2} + \lambda_2\mathcal{L}^{\mathcal{G}}_{adv}. \label{eq:generator}
\end{equation}
The discriminator is trained to judge whether its inputs are real or synthetic. The objective is to output $1$ when given input is the real data $\boldsymbol{x^{i}_{t+k}}$ and $0$ when input is generated data $\mathcal{G}(\boldsymbol{x^{i}_{t}})$. 
Therefore, the discriminator loss is defined as:
\begin{eqnarray}
\mathcal{L}^D = -log(\mathcal{D}(\boldsymbol{x^{i}_{t+k}})) - log(1-\mathcal{D}(\mathcal{G}(\boldsymbol{x^{i}_t})))\displaystyle.
\label{eq:discriminator}
\end{eqnarray}
\subsection{Action classifier and inference}
To evaluate the authentication of predicted features generated by the feature matching RNN, we again use the frame features to train a 2-layer MLP appended with a RBF kernel layer (equation~\ref{eq:ksvm}) to classify videos as early as possible. 
The classification loss is evaluated using a cross-entropy loss.
\Name and the action classification MLP is trained separately.
One might consider training both MLP and the \name jointly. 
However, in terms of performance, we did not see that much of advantage.
\begin{figure}[t]
\begin{minipage}{.465\textwidth}
      \centering
      \includegraphics[trim={0 0 2cm 0},clip,width =\textwidth]{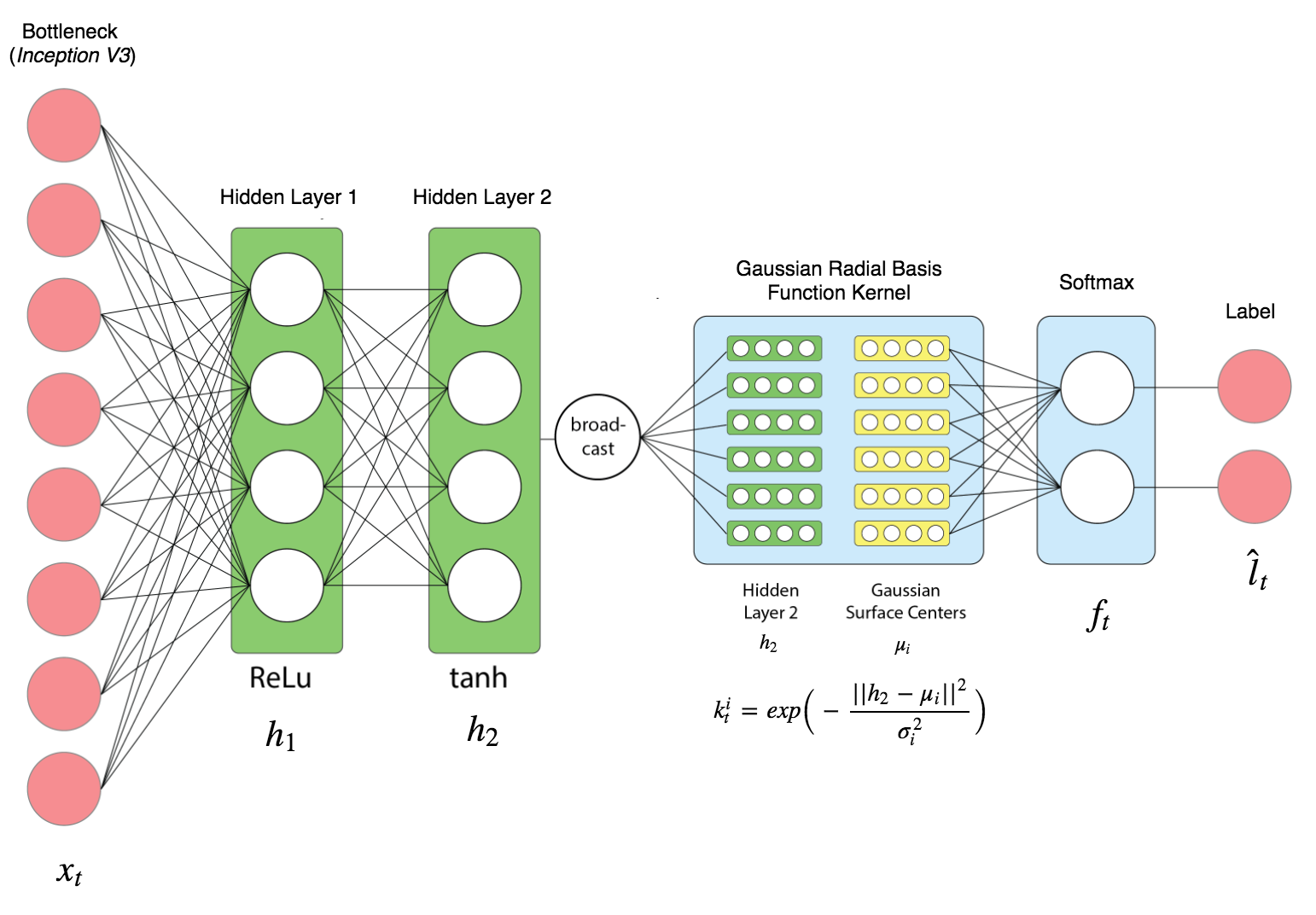}
      \caption{Illustration of RBF keneralized multilayer perceptron.}
      \label{fig:kMLP} 
\end{minipage}
\hspace{15pt}
\begin{minipage}{.465\textwidth}
\centering
\includegraphics[width = \textwidth]{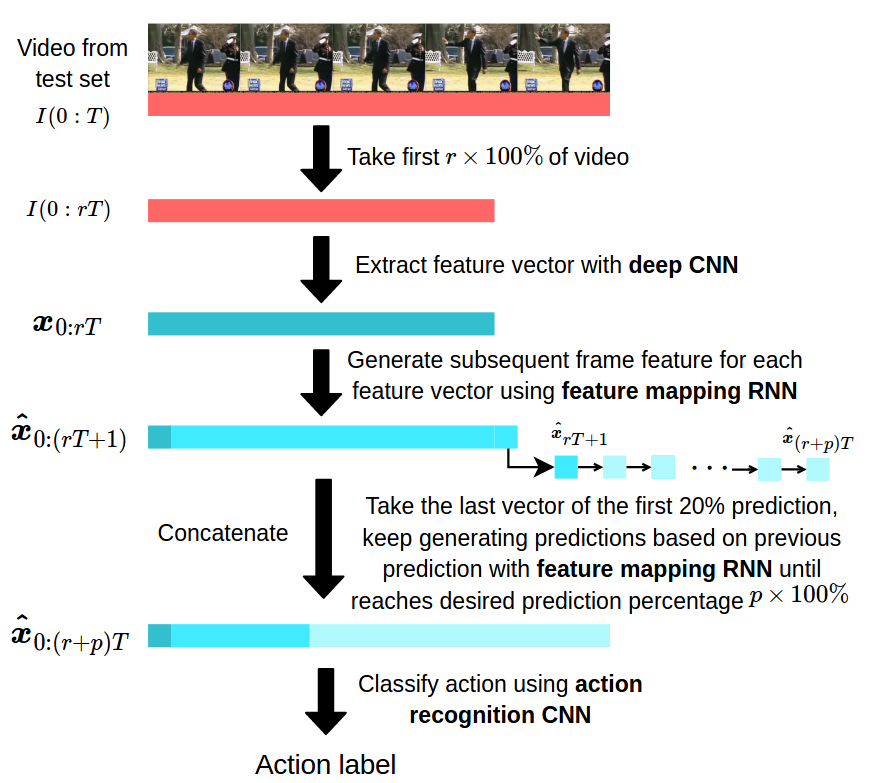}
\caption{Testing Procedure of \NAME} \label{fig:tp}
\end{minipage}
\end{figure}

During inference, we take advantage of all observed and generated features to increase the robustness of the results. Accuracy is calculated by performing temporal average pooling on all predictions (see ~Fig~\ref{fig:tp}).\par
\section{Experiments}\label{sec:experiments}
\subsection{Datasets}
Three datasets are used to evaluate the performance of our model, namely \textit{UT-Interaction} \cite{uti}, \textit{JHMDB-21} \cite{jhuanghJHMDB} and \textit{UCF101-24} \cite{UCF101}.
We follow the  standard protocols for each of the datasets in our experiments. 
We select these datasets because they are the most related to action anticipation task that has been used in prior work~\cite{AkbarianSSFPA17,msryoo2011}.\par
\header{UT-Interaction}
The \textit{UT-Interaction} dataset (UTI) is a popular human action recognition dataset with complicated dynamics. The dataset consists of 6 types of human interactions executed under different backgrounds, zoom rates and interference.
It has a total of 20 video sequences split into 2 sets.
Each video is of approximately 1 minute long, depicting 8 interactions on average. 
The available action classes include handshaking, pointing, hugging, pushing, kicking and punching. 
The performance evaluation methodology requires the recognition accuracy to be measured using a 10-fold leave-one-out cross validation per set. 
The accuracy is evaluated for 20 times while changing the test sequence repeatedly and final result is yielded by taking the average of all measurements. 
\par\header{{JHMDB-21}}
\textit{JHMDB-21} is another challenging dataset that contains 928 video clips of 21 types of human actions. 
Quite different from the \textit{UT-interaction} where video clips of different actions are scripted and shot in relatively noise-free environments, all videos in \textit{JHMDB-21} are collected from either movies or online sources, which makes the dataset a lot more realistic. 
Each video contains an execution of an action and the dataset is split into 3 sets for training, validation and testing.
\par\header{{UCF101-24}}
\textit{UCF101-24} is a subset of \textit{UCF101}. The dataset consists of more than 3000 videos from 24 action classes of \textit{UCF101}. Since all the videos are collected from YouTube, the diversity of data in terms of action types, backgrounds, camera motions, lighting conditions etc are guaranteed. In addition, each video depicts up to 12 actions of the same category with different temporal and spatial features, which makes it one of the most challenging dataset to date.
\subsection{Implementation Details}\label{sec:featurematchingrnn}
\header{\NAME} The \NAME is trained with batch size of $128$, using a hidden size ($H$) of 4 in all experiments unless otherwise specified. 
The default dimensionality of feature sub vector referred to as \emph{feature step size($D$)} is set to 128. 
We make use of six RBF kernels within the RBF kernelized \name. 
Feature stride is set to 64 and weight of the adversarial loss ($\lambda_1$) is set to 1 and the weight for $\mathcal{L}_2$ loss is set to 10 (\ie $\lambda_2$).
\par\header{Action classifier MLP} The a simple two layer MLP classifier consists of two hidden layers with 256 and 128 activation respectively. 
We also use RBF kernels along with the MLP where number of kernels set to 256. MLP is trained with batch size of $256$.
\par\header{Training and Testing Procedures}
We use pre-trained Inception V3~\cite{Szegedy2016} penultimate activation as the frame feature representation. 
The dimensions of each feature vector is 2048 ($d=2048$).
The action classification MLP is trained on the feature vectors from the training split of the datasets. 
These features are also used to train our \name to generate future features. 
Both models are trained with learning rate $0.001$ and exponential decay rate $0.9$. 
\par\header{Protocols} Following the experimental protocol~\cite{AkbarianSSFPA17,msryoo2011}, we used only the first $r\%$ ($50\%$ for \textit{UT-Interaction} and $20\%$ for \textit{JHMDB-21}) of the video frames to predict action class for each video. 
To utilise our model, we generate extra $p\%$ (referred to as prediction percentage) of the video features using our RBF kernalized \name. 
Therefore, we make use of $(r+p)\%$ feature vectors of the original video length to make the final prediction. 
To generate the next future feature at test time, we recursively apply our \name given all previous features (including the generated ones). 
We then use our action classification MLP to predict the action label using max pooling or simply average the predictions.
This procedure is demonstrated more intuitively in Fig.\ref{fig:tp}.
\subsection{Comparison to State-of-the-Art}
We compare our model to the state-of-the-art algorithms for action anticipation task on the \textit{JHMDB-21} dataset. 
Results are shown in Table~\ref{tab:sota-jhmdb}. Our best algorithm (denoted as \textit{fm+RBF+GAN+Inception V3} in the table) outperforms the state-of-the-art by $18\%$, and we can clearly see that the implementation of kernel SVM and adversarial training improves the accuracy by around $3$ to $4\%$. In addition, to show the progression of how our method is able to outperform the baseline by such a large margin, we also implemented the \NAME on top of VGG16 so that the deep CNN pre-processing is consistent with other methods in Table~\ref{tab:sota-jhmdb}. The \textit{fm+VGG16} entry in the table shows an $8\%$ improvement from baseline \textit{ELSTM}, which is purely influenced by the implementation of \NAME.

Experiments are also carried out on the two other mentioned datasets, where our best method outperforms the state-of-the-art by $13\%$ on \textit{UT-Interaction} and $6\%$ on \textit{UCF101-24}, as shown in Table~\ref{tab:sota-uti} and Table~\ref{tab:sota-ucf} respectively. 

We believe these significant improvements suggests the effectiveness of two main principles, the parameter sharing and expressive capacity of RBF functionals.
To further investigate the impact of each component, we perform a series of experiments in the following sections.
\begin{table}[t]
\begin{minipage}{.48\linewidth}\centering
\caption{Comparison of our model against state-of-the-arts on \textit{JHMDB-21} dataset for action anticipation. We follow the protocol of \textit{JHMDB-21} for action anticipation and predictions are made from using only $20\%$ of video sequence.}
\begin{tabular}{|c|l|c|}
\hline
&\textbf{Method} & \textbf{Accuracy} \\\hline
\multirow{6}{*}{\textbf{Others}}&ELSTM \cite{AkbarianSSFPA17} & $55\%$\\
&Within-class Loss \cite{shugaoma2016} & $33\%$\\
&DP-SVM \cite{soomro2016online} & $5\%$\\
&S-SVM \cite{soomro2016online} & $5\%$\\
&Where/What \cite{soomro2016predicting} & $10\%$\\
&Context-fusion \cite{jain2016} & $28\%$\\
\hline
\multirow{4}{*}{\textbf{Ours}}&fm+VGG16 & $\boldsymbol{63\%}$\\
&fm+kSVM+GAN+VGG16 & $\boldsymbol{67\%}$\\
&fm+Inception V3 & $\boldsymbol{70\%}$\\
&fm+RBF+GAN+Inception V3 & $\boldsymbol{73\%}$\\ \hline
\end{tabular}
\label{tab:sota-jhmdb}
\end{minipage}\hspace{20pt}
\begin{minipage}{.48\linewidth}\centering
\caption{Comparison of our model against state-of-the-arts on \textit{UT-Interaction} dataset for action anticipation. Following protocol of \textit{UT-Interaction}, predictions are made from using only $50\%$ of video sequence.}
\begin{tabular}{|l|c|}\hline
\textbf{Method} & \textbf{Accuracy} \\\hline
ELSTM \cite{AkbarianSSFPA17} & $84\%$\\
Within-class Loss \cite{shugaoma2016} & $48\%$\\
Context-fusion \cite{jain2016} & $45\%$\\
Cuboid Bayes \cite{msryoo2011} & $25\%$\\
I-BoW \cite{msryoo2011} & $65\%$\\
D-BoW \cite{msryoo2011} & $70\%$\\
Cuboid SVM \cite{ryoo2010overview} & $32\%$\\
BP-SVM \cite{laviers2009} & $65\%$\\
\hline
\textbf{Ours} & $\boldsymbol{97\%}$\\\hline
\end{tabular}
\label{tab:sota-uti}
\end{minipage}

\end{table}

\begin{table}[H]
\centering
\caption{Comparison of our model against state-of-the-arts on \textit{UCF101-24} dataset for action anticipation. Again, predictions are made from using only $50\%$ of video sequence.}
\begin{tabular}{|l|c|}\hline
\textbf{Method} & \textbf{Accuracy} \\\hline
Temporal Fusion \cite{fan2016online} & $86\%$\\
ROAD \cite{singh2016online} & $90\%$\\
ROAD + BroxFlow \cite{singh2016online} & $92\%$\\
\hline
\textbf{Ours} & $\boldsymbol{98\%}$\\\hline
\end{tabular}
\label{tab:sota-ucf}
\end{table}

\subsection{Analysis}\label{sec:modelcomparison}

In this section we compare the influence of different components of our RBF kernelized \name.
As shown in Table~\ref{tab:ac}, we compare following variants of our RNN model, including:
\vspace{-5pt}
\begin{enumerate}[leftmargin=0.15in]
\item[] {\textbf{(a) \NAME}: use only $\mathcal{L}_2$ loss to train the \NAME;}
\item[] {\textbf{(b) \NAME+RBF}: our RNN with kernalised RBF, still only using $\mathcal{L}_2$ loss;}
\item[] \textbf{(c) \NAME + RBF + GAN}: RBF kernelized feature mapping RNN with adversarial loss.
\end{enumerate}
\vspace{-5pt}
Apart from the \NAME-based models, we also conduct experiments on the following method as comparisons to our model:
\vspace{-5pt}
\begin{enumerate}[leftmargin=0.15in]
\item[] \textbf{(d) Linear}: a matrix of size $D \times D$ is used for feature generation ($D$ is dimension of input feature);
\item[] \textbf{(e) Vanilla LSTM}: generate future action features with traditional vanilla LSTM.  $\mathcal{L}_2$ loss is used to train it;
\item[] \textbf{(f) Vanilla LSTM + RBF}: vanilla LSTM with kernalised RBF, using only $\mathcal{L}_2$ loss;
\item[] \textbf{(g) Vanilla LSTM + RBF + GAN}: RBF kernalized vanilla LSTM with added adversarial loss.
\end{enumerate}
\vspace{-5pt}
Note that all the results are obtained using features extracted by Inception V3 network, and the accuracy are acquired using max pooling at prediction percentage $p=50\%$.
\vspace{-10pt}
\begin{table}
\centering
\caption{Comparison of different approach on \textit{JHMDB-21} dataset}
\begin{tabular}{|l|c|}\hline
\textbf{Method}&\textbf{Accuracy}\\ \hline
Linear&$62.7\%$\\
Vanilla LSTM&$66.3\%$\\
Vanilla LSTM + RBF&$67.9\%$  \\
Vanilla LSTM + RBF + GAN&{-}\\ \hline
Feature Mapping RNN&$72.2\%$\\
Feature Mapping RNN + RBF&$72.8\%$\\
Feature Mapping RNN + RBF + GAN&$73.4\%$\\  \hline
\end{tabular}
\label{tab:ac}
\end{table}

The results in Table~\ref{tab:ac} shows the proposed scheme outperforms the linear model significantly while using fewer parameters.
Most interestingly, the \name outperforms vanilla LSTM by almost 6\% indicating the impact of parameter sharing in the feature space.
We can also conclude from Table~\ref{tab:ac} that the application of adversarial loss as well as RBF kernel layers encourages the model to generate more realistic future features, which is reflected by the improvement in accuracy with \NAME+RBF and \NAME+RBF+GAN.
It is also shown in the Table \ref{tab:ac} that vanilla LSTM trained with RBF kernel yields almost $2\%$ higher accuracy than plain vanilla LSTM, which proves further that the RBF layer is something the baseline can benefit from. 
Regrettably, the vanilla LSTM with adversarial training model failed to stabilise due to large number of parameters needed in the LSTM cells to reconstruct the original feature distribution.

The influence of RBF kernalized \name is quite distinctive. If we compare the red curve to the green one, we can see that the discrepency between them becomes larger as the prediction percentage increases. This indicates that the RBF kernalized \name generate more accurate future features in the long term, and hence it is a more robust model than plain \name. Comparing the red and green curve to the orange and blue one, we can also conclude that the adversarial loss assist the RNN training in a similar way. Even without the assistance of GAN loss and RBF kernel, the \name still performs better than liner projection RNN.

\begin{figure}[t]
\begin{minipage}{.48\textwidth}
        \centering
        \includegraphics[width = \textwidth]{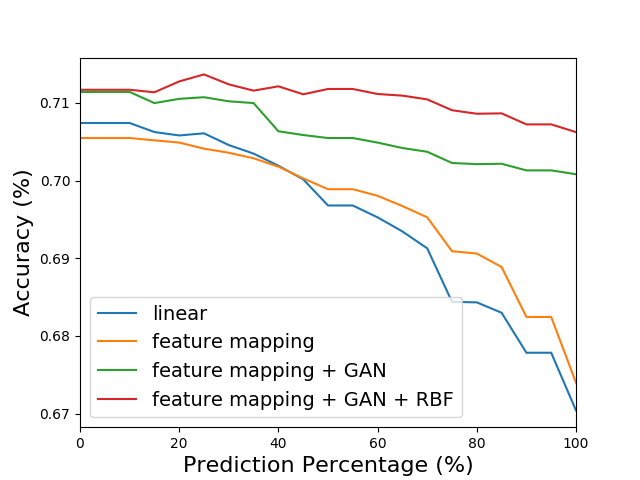}
    \caption{Prediction accuracy without pooling for JHMDB-21 dataset at different video prediction percentages $\boldsymbol{p}$. RBF kernalized \Name is trained using adversarial loss is able to achieve the highest stable accuracy.}
    \label{fig:accuracyvp}
\end{minipage}
\hspace{15pt}
\begin{minipage}{.48\textwidth}
      \centering
        \includegraphics[width = \textwidth]{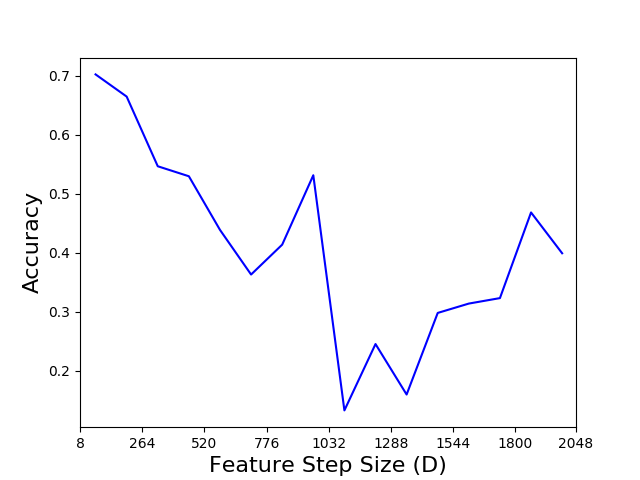}
    \caption{Prediction accuracy evaluated at different feature step sizes on \textit{JHMDB-21} dataset. The accuracy plotted in the image is found by implementing feature step size between $D=8$ to $2048$ with increment of $8$ on the model and the rolling average is taken among every $16$ measurements. No temporal pooling is used.}
    \label{fig:pafss}
\end{minipage}
\end{figure}

\subsection{Influence of Hyper-parameters}\label{sec:featurestepsize}
\header{Feature Step Size}
The accuracy of the generated data indicates the existence of strong correlations between the $D$-dimensional segments of the feature vectors. 
By default, we resort to feature step size of 128 ($D=128$).
In order to further explore this property, we experimented with different feature step sizes.
\begin{table}[t]
\small
\begin{minipage}{.3\linewidth}
\centering
\caption{Prediction accuracy at different feature stride size ($\boldsymbol{S}$)}
\begin{tabular}{|c|c|}\hline
Interval Size&Accuracy\\ \hline
$S=4 $&$74.3\%$\\
$S=8 $&$73.8\%$\\
$S=16 $&$74.3\%$\\
$S=32 $&$73.2\%$\\
$S=64 $&$73.2\%$\\
$S=128$&$72.4\%$\\ \hline
\end{tabular}
\label{tab:stride}
\end{minipage} \hspace{12pt}
\begin{minipage}{.3\linewidth}
\centering
\caption{Prediction accuracy using LSTM cells with different state size ($\boldsymbol{H}$).} 
\begin{tabular}{|c|c|}\hline
Hidden State Size&Accuracy\\ \hline
$H=2 $&$71.7\%$\\
$H=4 $&$73.2\%$\\
$H=8 $&$72.7\%$\\
$H=16$&$73.2\%$\\ 
$H=32$&$73.2\%$\\
$H=64$&$73.8\%$\\ 
\hline
\end{tabular}
\label{tab:ivapkd}
\end{minipage}\hspace{12pt}
\begin{minipage}{.3\linewidth}
\centering
\caption{Prediction accuracy using different number of RBF kernels.}
\begin{tabular}{|c|c|}\hline
No. of Kernels &Accuracy\\ \hline
$k=4  $&$72.7\%$\\
$k=8  $&$72.7\%$\\
$k=16 $&$73.3\%$\\ 
$k=32 $&$73.3\%$\\
$k=64 $&$72.7\%$\\ 
$k=128$&$73.8\%$\\ 
$k=256$&$72.2\%$\\ 
\hline
\end{tabular}
\label{tab:inrk}
\end{minipage}
\end{table}
In Fig.\ref{fig:pafss}, we plot the recognition accuracy against feature step size. 
We observe that small feature step size guarantees effective feature generation. 
Specifically, the prediction remains above $70\%$ when feature step size is smaller than $200$. 
This phenomena can be explained by the intuition that when feature step size is large, the model tries to generalize a large set of features with mixed information at one time step, which results in degraded performance.

It is also interesting to note that the prediction accuracy oscillates drastically as the feature step size exceeds $250$. 
This indicates that perhaps the feature vector summarizes information of the original image in fixed-size clusters, and when we attempt to break these clusters by setting different feature step size, the information within each time step lacks continuity and consistency, which subsequently compromises the prediction performance.\par
Although smaller feature step size builds a more robust model, the training time with feature step size $16$ takes only half the amount of time of training with step size $4$, with no compromise on prediction accuracy. Therefore, it might be beneficial sometimes to choose a larger feature step size to save computational time.
\par\header{Interval Size}
In this section we experiment the effect of overlapping sub-feature vectors on our RBF kernalized \name. 
Recall that the \name is denoted by $\mathcal{G}: \boldsymbol{x_t^{i}}\rightarrow \boldsymbol{\hat{x}_{t+k}^{i:}}$. 
Instead of incriminating $i$ by the multiple of feature step size $D$, in an attempt to improve the prediction accuracy, we define an feature stride $S$ that is smaller than $D$. The prediction accuracy of \NAME with several different feature stride value is shown in Table~\ref{tab:stride}.
\par\header{LSTM state size}
This section aims at investigating the influence of LSTM cell's hidden state size ($H$) on the model's performance. 
Since the hidden state stores essential information of all the input sequence data, it is common to consider it as the "memory" of the RNN. 
It is intuitive to expect an improvement in performance when we increase the size of the hidden state up to some extent.\par 
However, the results in Table~\ref{tab:ivapkd} shows that increasing the LSTM state size does not have much effect on the prediction accuracy, especially when the state size becomes larger than $8$. 
This is because in the proposed \name model, each LSTM cell takes only one scalar as input, as opposed to the traditional RNN cells that process entire vectors. 
As the hidden state size is always greater than the input size (equal to 1), it is not surprising that very large $H$ does not have much influence on the model performance.
\par\header{Number of RBF Kernels}
In this section we study the influence of number of Gaussian surfaces used in \name. 
We calculate prediction accuracy while increasing the number of Gaussian kernels from $2^1$ to $2^8$. Results are as shown in Table~\ref{tab:inrk}.
The results show a general trend of increasing prediction performance as we add more number of kernels, with the highest accuracy achieved at when $k=128$. 
However, result obtained when $k=256$ is worse than when $k=4$. This phenomena could be explained by over-fitting, resulted from
 RBF kernel's strong capability of modeling temporal dynamics of data with complex boundaries. 
\par\header{Conclusions for hyper-parameters tuning} 
The conclusion from these experiments is that the model is not too sensitive to the variation of these hyper-parameters in general, which demonstrates  its robustness. Results further demonstrated the computational efficiency of our approach. Since it is possible to effectively train the model with very few parameters, it can be stored on mobile devices for fast future action anticipation.

\section{Conclusions}

The proposed RNN which uses a very few parameters outperforms state-of-the-art algorithms on action anticipation task. 
Our extensive experiments indicates the model's ability to produce accurate prediction of future features only observing a fraction of the features.
Furthermore, our RNN model is fast and consumes fraction of the memory which makes it suitable for real-time execution on mobile devices. 
Proposed \name can be trained with and without lables to generate future features.
Our feature generator does not use class level annotations of video data. Therefore, in principle, we can increase the robustness of the model utilizing large amount of available unlabelled data. 
The fact that the model is able to generate valid results using very few parameters provides strong proofs for the existence of inner-correlation between deep features, which is a characteristic that can have implications on many related problems such as video tracking, image translation, and metric learning.\par
In addition, by appending a RBF layer to the RNN, we observe significant improvement in prediction accuracy. 
However, it was also noted that over-fitting occurs when the model is implemented with too many kernel RBFs. 
To fully explore functional capacity of RBF function, in future studies, we aim to implement kernel RBFs on fully connected layer of popular deep CNN models such as \textit{ResNet} \cite{2015kaimingheRESENET}, \textit{AlexNet} \cite{alexkrizhevsky2012} and \textit{DenseNet} \cite{huanggao2016DENSENET}. 

In conclusion, proposed RBF kernalized \name demonstrates the power of parameter sharing and RBF functions in a challenging sequence learning task of video action anticipation.

\bibliographystyle{splncs04}
\bibliography{bibliography}

\end{document}